\documentclass[conference]{IEEETran}
\IEEEoverridecommandlockouts
\usepackage{cite}
\usepackage{amsmath,amssymb,amsfonts}
\usepackage{algorithmic}
\usepackage{booktabs}
\usepackage{graphicx}
\usepackage{tabularx}
\usepackage[font=small]{caption}
\usepackage{textcomp}
\usepackage{xcolor}
\def\BibTeX{{\rm B\kern-.05em{\sc i\kern-.025em b}\kern-.08em
    T\kern-.1667em\lower.7ex\hbox{E}\kern-.125emX}}
    
\graphicspath{{./figures/}}

\begin{document}

\title{Sparse Mutation Decompositions: Fine Tuning Deep Neural Networks with Subspace Evolution}

\author{\IEEEauthorblockN{Tim Whitaker}
\IEEEauthorblockA{\textit{Department of Computer Science} \\
\textit{Colorado State University}\\
Fort Collins, CO, USA \\
timothy.whitaker@colostate.edu \\[-3.0ex]}
\and
\IEEEauthorblockN{Darrell Whitley}
\IEEEauthorblockA{\textit{Department of Computer Science} \\
\textit{Colorado State University}\\
Fort Collins, CO, USA \\
whitley@cs.colostate.edu \\[-3.0ex]}
}
\maketitle

\begin{abstract}
Neuroevolution is a promising area of research that combines evolutionary algorithms with neural networks.
A popular subclass of neuroevolutionary methods, called evolution strategies, relies on dense noise perturbations to mutate networks, which can be sample inefficient and challenging for large models with millions of parameters.
We introduce an approach to alleviating this problem by decomposing dense mutations into low-dimensional subspaces.
Restricting mutations in this way can significantly reduce variance as networks can handle stronger perturbations while maintaining performance, which enables a more controlled and targeted evolution of deep networks.
This approach is uniquely effective for the task of fine tuning pre-trained models, which is an increasingly valuable area of research as networks continue to scale in size and open source models become more widely available.
Furthermore, we show how this work naturally connects to ensemble learning where sparse mutations encourage diversity among children such that their combined predictions can reliably improve performance.
We conduct the first large scale exploration of neuroevolutionary fine tuning and ensembling on the notoriously difficult ImageNet dataset, where we see small generalization improvements with only a single evolutionary generation using nearly a dozen different deep neural network architectures.

\end{abstract}


\section{Introduction}

\begin{figure*}[t]
    \centering
    \includegraphics[width=2\columnwidth]{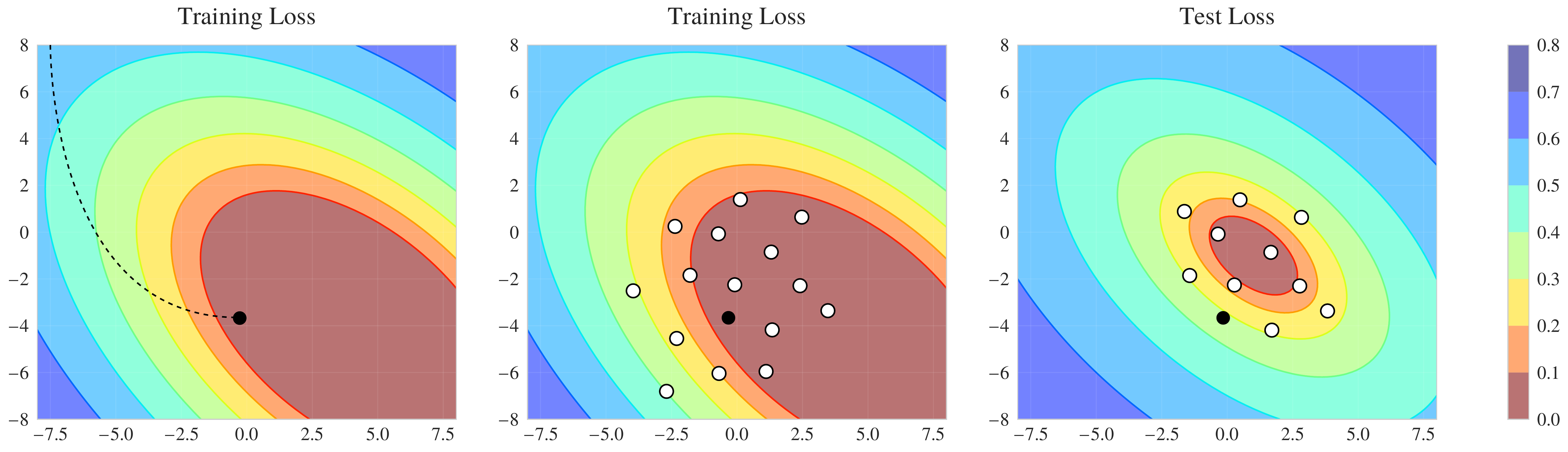}
    \caption{Neuroevolution excels at fine tuning fully-trained networks that get stuck in flat loss basins. The left graph displays a typical SGD training trajectory where models tend to converge to the edges of flat optima \cite{https://doi.org/10.48550/arxiv.1803.05407}. The middle graph displays white dots as child networks that are generated from sparse mutations. The right graph shows the final ensemble consisting of top candidates selected from evaluation on a separate validation set. A key insight to the success of gradient-free methods for fine tuning is that the loss landscapes of test distributions rarely match the training distribution exactly. Ensembling over wide areas of good validation performance is key to improving generalization.}
    \label{fig:my_label}
\end{figure*}

Neuroevolutionary methods evolve populations of models using biologically inspired concepts like natural selection and mutation. A core component of many of these methods is the generation of new networks through random noise perturbations. However, as neural networks continue to grow in size, the effect of noise perturbations on network behavior becomes increasingly pronounced. There is often a critical region of mutation where too little of a change does not provide any meaningful exploration, and too significant of a change leads to complete performance collapse. The optimal mutation window can be vanishingly small with modern deep neural networks that contain millions or billions of parameters and different network architectures, layer configurations, and training optimizers can lead to significant differences in parameter sensitivity. Investigating techniques to make mutations more effective and tractable for deep neural networks is important for opening new avenues of neuroevolutionary research.

Due to the difficulty in mutating these large models, neuroevolutionary methods have primarily seen success with tasks that can be solved by relatively small networks \cite{Galvan_2021}.
This is compounded by the fact that these methods tend to be sample inefficient on supervised learning problems with well defined gradients.
Hybrid gradient/evolutionary methods have thus grown in popularity as gradient optimization can be used to rapidly train the model while evolutionary processes can be implemented to aid in exploration and fine grained convergence \cite{DEBRUIN20208049}.
This idea naturally extends to the task of fine tuning and optimizing large models that have been pre-trained with gradient descent.
This is especially important as networks and labeled datasets continue to scale in size which can make training from scratch prohibitively expensive.
Pre-trained and open source models are becoming more widely available, which makes methods for improving them increasingly valuable.


In order to alleviate the challenges of mutating deep networks, we introduce Sparse Mutation Decompositions as a method for breaking up dense mutations into low-dimensional subspaces.
While sparse mutations have long been used in other areas of evolutionary and genetic programming, they have rarely been explored with the popular evolution strategy based approaches \cite{evostrat, https://doi.org/10.1002/fedr.19750860506, https://doi.org/10.48550/arxiv.1604.00772, NEURIPS2018_85fc37b1}.
This is likely due to the perceived sample inefficiency associated with only updating small numbers of parameters at a time, however we find that this can actually be desirable for the task of mutating pre-trained models.
Reducing the dimensionality of noise perturbations widens the critical mutation window, which significantly reduces variance as networks can handle stronger perturbations before performance collapse.
We also explore the notion of static and dynamic subspace evolution in which mutations are restricted to the same or different subspaces for each child.
Our work naturally connects to ensemble learning where we explore how stronger but more sparse mutations can encourage diversity among children such that the combined predictions of a mutated population can reliably improve generalization performance.

Along with several ablation studies and visualizations designed to gain insight into the interplay between mutation strength and sparsity, we introduce the first large scale exploration of neuroevolutionary fine tuning with sparse mutations on the notoriously difficult ImageNet dataset with nearly a dozen deep convolutional network architectures. Using ensembles of mutated populations results in monotonic generalization improvements of up to 0.5\% using only a single evolutionary generation and with no additional training.

\section{Background}

\subsection{Neuroevolution}

Neuroevolution has long been a promising area of research in the machine learning community as evolutionary algorithms offer an elegant approach to optimizing neural networks by utilizing natural and biological metaphors like natural selection, mutation, genetics and reproduction \cite{WHITLEY1990347}.
These methods typically employ generational loops where populations of members are spawned from some parent(s) using evolutionary operators and evaluated for their fitness on a validation set.
The best models are then selected and used as parents for the next generation.

Evolution Strategies (ES) are the most popular subclass of neuroevolutionary methods that use only selection and noise mutations to optimize weights \cite{https://doi.org/10.1002/fedr.19750860506}. These methods track the mean and standard deviation of parameters modeled by a Gaussian distribution and offspring are generated by sampling from this distribution. The subset with the highest fitness scores are selected and the distribution's mean and standard deviation are updated according to the parameters of the elite subset. Recent work from Open AI has shown that a wide scaling of simple evolution strategies can be incredibly powerful on difficult reinforcement learning tasks \cite{https://doi.org/10.48550/arxiv.1703.03864}.

Covariance Matrix Adaptation Evolution Strategies (CMA-ES) improves on ES by tracking dependencies between member parameters in a population with a pairwise covariance matrix. The standard deviation can then control for the overall scale of the distribution, greatly improving exploration \cite{https://doi.org/10.48550/arxiv.1604.00772}. However, despite the exploratory power of this approach, constructing the pairwise covariance matrix is computationally expensive and impractical for large networks.

These approaches to optimization have several advantages over gradient based methods that are typically used in deep learning.
They are inherently scalable as population members can be evaluated independently and they excel at exploring landscapes where gradient information is noisy, flat, or unavailable \cite{https://doi.org/10.48550/arxiv.1703.03864}.
This is common in reinforcement learning environments where reward information can be sparse or dynamic.
Unfortunately, evolutionary methods tend to be very sample inefficient on supervised learning problems.

Hybrid gradient/evolutionary methods are able to take advantage of the efficiency of gradient methods with the exploratory power of evolutionary methods \cite{NEURIPS2018_85fc37b1, hybrid}.
Sparse Evolutionary Training is one related approach that uses alternating phases where evolutionary algorithms are used to determine which subnetwork to train during a given phase with gradient descent \cite{Mocanu_2018}.

Neuroevolutionary methods are also known for exploring the topological space of networks \cite{stanley2002evolving}.
In these methods, the network architecture itself is evolved as well as the weights.
This can be an effective way to encourage diversity among population members as different network structures force unique representations through their structure.
Sparse mutations hold an interesting connection to this line of work as there is growing interest into the nature of subnetwork behavior in trained models \cite{https://doi.org/10.48550/arxiv.1701.08734, https://doi.org/10.48550/arxiv.1803.03635}.

Several researcher have noted the natural connection between evolutionary populations and ensemble learning \cite{https://doi.org/10.48550/arxiv.0704.3905, 887237, 4442253}.
Ensembles of diverse and accurate models consistently improve upon generalization performance as using multiple predictions can help to reduce the bias or variance associated with making predictions using a single model \cite{brown2005diversity, 10.1007/3-540-45014-9_1}.

\subsection{Model Tuning}

Much of deep learning optimization research is focused on methods for training neural networks well from scratch.
However, as neural networks continue to scale in size, the cost for training these large networks from scratch becomes expensive.
As open source and pre-trained models are becoming more widely available, investigating how we can take trained models and improve them further becomes a valuable area of research.


Transfer learning is one line of work that has popularized the importance of utilizing pre-trained model. This generalized the notion of using a network trained on one task or dataset (usually a much larger and general set like ImageNet) and then tuning it on another \cite{inbook, bozinovski2020reminder}.
Since the original model is fully trained, the model tends to converge much quicker on the new task than it would have taken if trained from scratch.
Several researchers have also explored freezing early weights in the network during the tuning phase, as early layers in deep networks tend to learn general patterns and features that don't necessarily need to be retrained \cite{https://doi.org/10.48550/arxiv.1911.02685}.
This is naturally connected to our approach where large numbers of parameters are kept frozen in order to maintain behavior while sparse subnetworks are mutated.


We generalize the term fine tuning to refer to the continued training of any optimized model. It is most popularly connected to network pruning literature that refers to the final process of training a sparse network after parameters are removed from a dense network.
Pruning and fine tuning is a well known technique for compressing model sizes where networks can be made significantly smaller with little to no loss in generalization performance \cite{https://doi.org/10.48550/arxiv.2003.03033, https://doi.org/10.48550/arxiv.2202.02643, NIPS1989_6c9882bb}.
These small subnetworks hold a signficant amount of classification power which suggest that limiting mutations to sparse subnetworks can have a meaningful impact on network behavior.

Recent optimization research has investigated the convergence behavior of deep networks in the final epochs of optimization.
It is thought that neural networks that converge to wide and flat optima in the optimization landscape tend to generalize better than minima that are sharp \cite{https://doi.org/10.48550/arxiv.2002.03495}.
With gradient descent, deep networks generally converge to the edges of these wide and flat optima.
However, the edges of these minima rarely correspond to the minima of the test distribution which is more likely to exist somewhere in the middle of these optima. Since these optima are wide and flat, there is little gradient information that can be used to nudge the model towards those locations.

Stochastic Weight Averaging (SWA) is a method that leverages this insight to fine tune models in the final epochs of training \cite{https://doi.org/10.48550/arxiv.1803.05407}. By using a repeating cyclic schedule, large learning rates are used to jump the model to new locations around the minima before converging with small learning rates. This is repeated several times where the weights of the model at each convergence location are saved and eventually averaged together.

Model Soups operate by using a very similar principle to Stochastic Weight Averaging \cite{https://doi.org/10.48550/arxiv.2203.05482}.
Instead of using a single model and a repeating cyclic learning rate schedule, Model Soups instead train a pseudo-ensemble where they fine-tune several clones of a trained model, each with a unique set of hyperparameters. The fine-tuned models tend to converge to the same loss basin but in unique locations, and the weights of each model are then averaged together as in Stochastic Weight Averaging.

Both Stochastic Weight Averaging and Models Soups provide empirical evidence for the theoretical ideas behind our approach with neuroevolutionary tuning. Figure 1 illustrates this with an example about how evolution can be used effectively in situations where models trained with gradient descent get stuck in some flat minima. The generation of child networks using sparse mutations provides enough diversity to explore the local landscape where ensembling over these wide areas of good validation performance is effective for reliably increasing generalization.

\section{Sparse Mutation Decompositions}

Neuroevolutionary methods typically generate child networks by applying dense noise perturbations to all the weights of a given parent network.
Some methods have suggested rescaling mutations according to weight magnitudes or output gradients \cite{https://doi.org/10.48550/arxiv.1712.06563}.
Sparse Mutation Decompositions instead restrict mutations to a small subset of parameters, which can allow for a more targeted and controlled evolution of deep neural networks.
This can greatly widen the critical mutation range, in that we can apply stronger mutations to achieve more meaningful diversity before experiencing performance collapse that dense mutations would cause.
The methods introduced here are general and applicable to a wide variety of network architectures and optimization algorithms.

\subsection{Noise Perturbations}

Mutations are usually implemented by perturbing the weights of a parent model $\theta \in \mathbb{R}^w$, where $w$ is the number of weights, with a random noise vector $N \in \mathbb{R}^w$, sampled from a Gaussian distribution $\mathcal{N}(\sigma, \mu)$. The sparse mutation $\gamma$ is implemented by taking the Hadamard product $\circ$ of a binary bitmask $M \in \mathbb{B}^w$ with the sampled noise vector $N$.
\begin{align*}
M &= \{1,0\}^{w} \\
N &\sim \mathcal{N}_w(\mu, \sigma^2) \\  
\gamma &= N \circ M
\end{align*}

When generating a population of children, there is a distinction to be made between whether mutations are restricted to the same subspace for each child or mutations are applied to random subspaces for each child. We call these two approaches Static and Dynamic Subspace evolution. Restricting all children to a single subspace may be more efficient for converging to an optima, while having each child mutate a unique subspace may lead to better diversity and exploration. We explore the differences between these two methods in our mutation ablation experiments in section IV. In the context of fine tuning pre-trained models, there is little difference between static and dynamic subspaces. More research could investigate the efficacy of these two approaches with models that are trained from scratch.

\subsection{Hyperparameter Search}

A significant challenge in these approaches is to determine the appropriate values for both the noise distribution as well as the amount of sparsity in the bitmask.
Unfortunately, this is challenging to determine before hand due to significant differences between the size and complexity of datasets, model architectures, optimization hyperparameters, etc.

One approach is to measure the differences between the outputs of a network before and after mutations are applied. For example, given a network $F$ that is parameterized by $\theta$ with $O$ outputs and a network that is perturbed with a noise mutation $\gamma$, the mean squared difference over a set of $N$ total input samples $X$ can be described as:
\[
MSE = \frac{1}{N} \sum_{n=1}^N \sum_{o=1}^O (F(X_n; \theta)_o - F(X_n; \theta + \gamma)_o)^2
\]

However, since classification networks are often trained with cross entropy loss where the outputs are treated as a probability distribution, mean squared error may not be the best fit for approximating the effect that perturbation might have due to potential variance. In this case, we suggest that Kullback-Leibler Divergence, or relative entropy, is a better fit for approximately measuring how different the two output distributions are.
\[
KL = \frac{1}{N} \sum_{n=1}^N F(X_n; \theta) \log \left( \frac{F(X_n; \theta)}{F(X_n; \theta+\gamma)} \right)
\]

This approach then allows for a general measure of comparison that works regardless of network architecture or noise distribution. A simple search algorithm can then try out several values for both mutation strength and sparsity in order to target a divergence score that prioritizes either exploration or accuracy. One nice property revealed in our experiments is that there is a roughly linear relationship between KL divergence and accuracy degradation in trained models. This insight can be used to tweak mutation parameters according to problem complexity, population size, or parameter sensitivity where a larger KL target can allow for stronger mutations and better exploration at the cost of potentially less accurate candidate models.

\subsection{Anti-Random and Mirrored Noise}

Neuroevolutionary methods typically use large populations of small networks that can be evaluated very quickly.
Deep neural networks can be very costly in both runtime and memory requirements, which invariably means that our populations will be much smaller.
For this reason, the random perturbations can be a large source of variance since we have a much smaller pool of candidates to choose from.
Anti-random and mirrored noise can be effective tools for reducing this variance by automatically generating sets of opposed child networks.

In the case of dynamic subspace evolution, anti-random sampling can be used to maximize subspace distance between population members.
For a given network with $w$ weights and a random bit mask $M = \{0,1\}^w$, the most distant vector is one in which the polarity of all of the bits are flipped, $M' = 1 - M$ \cite{malaiya1995antirandom, shen2008antirandom}.
The two masks can then be applied to two different networks, resulting in an even exploration over all the parameters of the network.
This can be extended to multiple members in a population by instead partitioning the parameter space, where a group of $N$ bitmasks are created such that the children form a disjoint union over all parameters of the parent network.
Each child then mutates a unique set of parameters that are not shared by other networks.

Mirrored sampling is another effective technique for reducing variance that is common in evolutionary literature \cite{inproceedings, https://doi.org/10.48550/arxiv.1703.03864}. In this case, the magnitude of the noise vectors are flipped, such that for a given mutation vector $\gamma$, the mirrored noise would then be $-\gamma$.

For example, assume that a bit mask $M$ is randomly generated for a network containing $w$ weights by sampling a binary distribution with some probability $\rho$ that a parameter will be masked, and a noise vector $N$ is randomly generated by sampling from a Gaussian distribution. The four resulting child networks $C$, parameterized by $\theta$, can then be described:
\begin{align*}
M &\sim \text{Bernoulli}_w(\rho) \\
N &\sim \mathcal{N}_w(\mu, \sigma^2) \\
C_1 &= \theta + (N \circ M) \\
C_2 &= \theta + (N \circ (1 - M)) \\
C_3 &= \theta - (N \circ M) \\
C_4 &= \theta - (N \circ (1 - M))
\end{align*}

With these two techniques, a set of child networks can be automatically generated for every noise perturbation, resulting in a population that is more evenly distributed in space. 

\subsection{Predictions}

\begin{figure*}
    \centering
    \includegraphics[width=\textwidth]{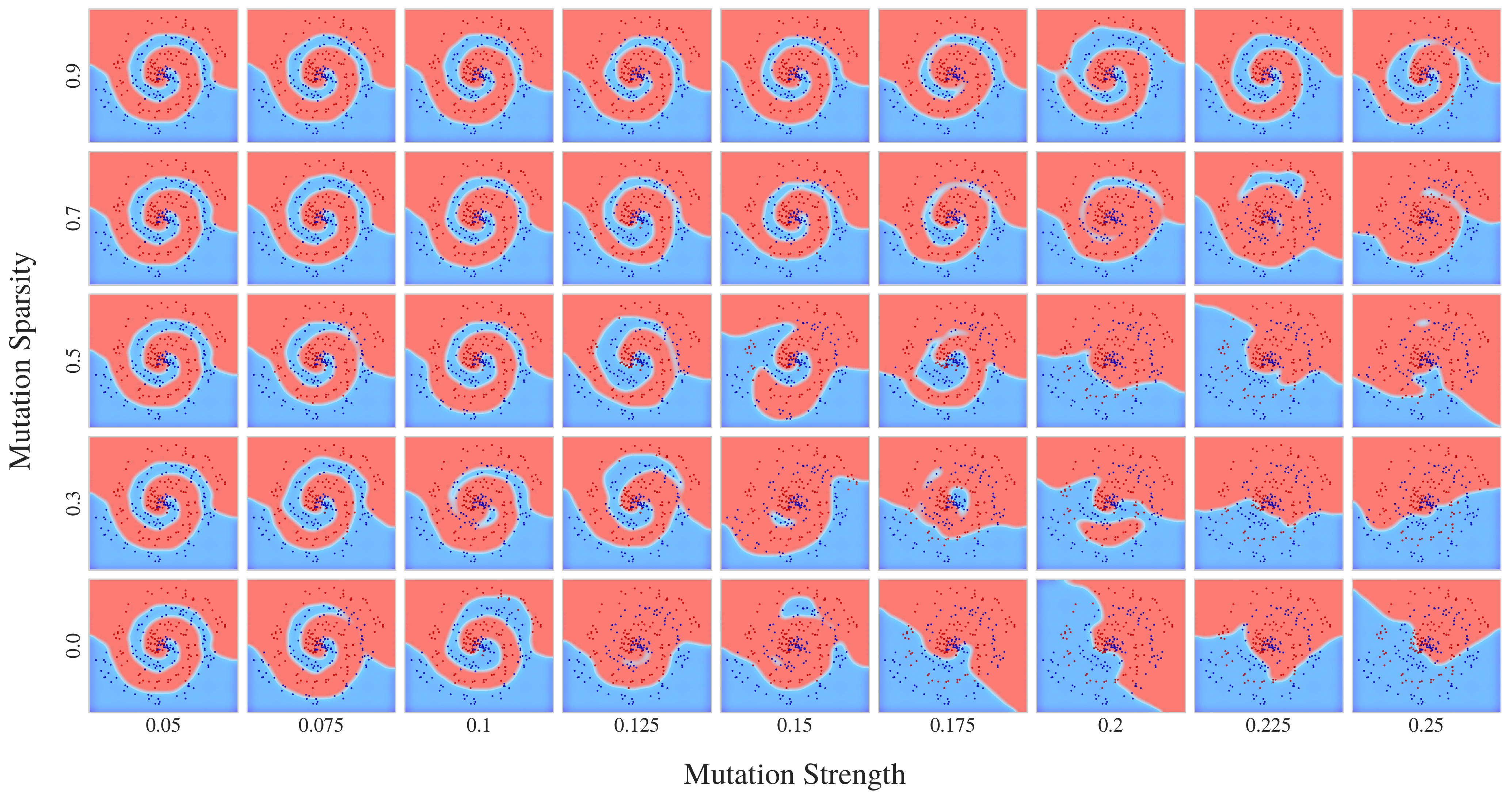}
    \caption{The images above detail decision boundaries for a trained three layer multilayer perceptron after being perturbed with various mutations. With dense perturbations, there is a comparatively small window between a functional decision boundary and complete performance collapse. As mutations become more sparse, the model is better able to retain its behavior while the strength of parameter mutation increases.}
\end{figure*}

Once we have a collection of mutated child networks, we evaluate each child on a validation set where the accuracy is recorded as a measure of its fitness.
Any model selection is done on this validation set, which is separate than the holdout test set we evaluate our final system on.
We then select child networks with the best fitness to be used for making predictions.

With a population of top candidates selected, we then need to combine them in order to make our final predictions.
The traditional evolutionary approach is to average the weights of the top candidates together into a single model.
There is also a natural connection to ensemble learning where the population can be evaluated independently and the predictions of each model can be combined.
We explore both approaches in our ablation experiment in section IV, where we find that averaged models maintain their performance for much stronger levels of mutation, while the generalization accuracy of ensembles tend to outperform averaged models for tuned mutation hyperparameters.

There are a lot of approaches for combining models predictions in ensembles, including majority vote, weighted model averaging, and bucket-of-models selection \cite{fragoso2017bayesian, dzeroski2004combining, vanderlaan2007super}.
For the purpose of this work, we use simple non-weighted prediction averaging which is standard practice for modern ensemble methods.

Mutations can affect the magnitude of raw outputs, so in order to achieve better normalization, we average the softmax of each ensemble member's outputs to ensure that all ensemble members produce outputs of the same scale.
\[ y_e = argmax( \frac{1}{S} \sum_{i=1}^{S} \sigma(f_i(x))) \]

where $y_e$ is the ensemble prediction, $S$ is the number of members in the ensemble (the ensemble size), $\sigma$ is the softmax function and $f_i(x)$ is the output of the individual ensemble member $i$.

\section{Experiments}

We first attempt to gain insight into how mutations influence network predictions by visualizing the decision boundaries of a trained network before and after mutating it. 
We then conduct a larger scale ablation experiment with a wide residual network where we generate populations of mutated networks with varying levels of mutation strengths and sparsities. We explore the differences between static and dynamic subspace evolution as well as differences between treating the top candidates in a population as an ensemble or averaging their weights together into one model. 
We end with a large scale experiment on the difficult ImageNet dataset with a dozen different model architectures.

\subsection{Decision Boundaries}

We begin by exploring the interplay between mutation sparsity and mutation strength by visualizing how changing these values affects network predictions.
We use a simple three layer fully connected multilayer perceptron that contains 64 neurons in each layer.
We train this model on a binary interleaved spiral dataset that contains 2500 sample (x, y) points. 
We train for 10 epochs using the Adam optimizer \cite{kingma2017adam} with a learning rate $0.001$ and use this trained model as the starting point for all mutations.

We then perturb the model with mutations sampled from a Gaussian Distribution $N \sim \mathcal{N}(0, \sigma^2)$ where $\sigma$ corresponds to the mutation strength and with a mask sampled from a Bernoulli distribution $M \sim \text{Bernoulli}(\rho)$ where $\rho$ governs the probability of mutation sparsity. We ablate these hyperparameters from mutation strengths of $\sigma \in [0.05, 0.25]$ and mutation sparsities of $\rho \in [0.0, 0.9]$.

Using the perturbed models, we make predictions on a holdout test set containing 250 samples. In figure 2, we display a grid of the decision boundaries of these perturbed models on the test set as we ablate between mutation strength and sparsity.
When mutations are dense, we see a very quick collapse of performance for small perturbations. The window for optimal dense mutation is quite small even for this toy network and simple dataset. When mutations become more sparse, the model is able to maintain a good classification boundaries while displaying small variations of prediction diversity. This kind of behavior is desirable for neuroevolutionary populations as ensemble learning research has shown that populations perform better with large numbers of both accurate and diverse members \cite{bonab2016theoretical}.

\subsection{Mutation Ablations}

\begin{figure*}
    \centering
    \includegraphics[width=\textwidth]{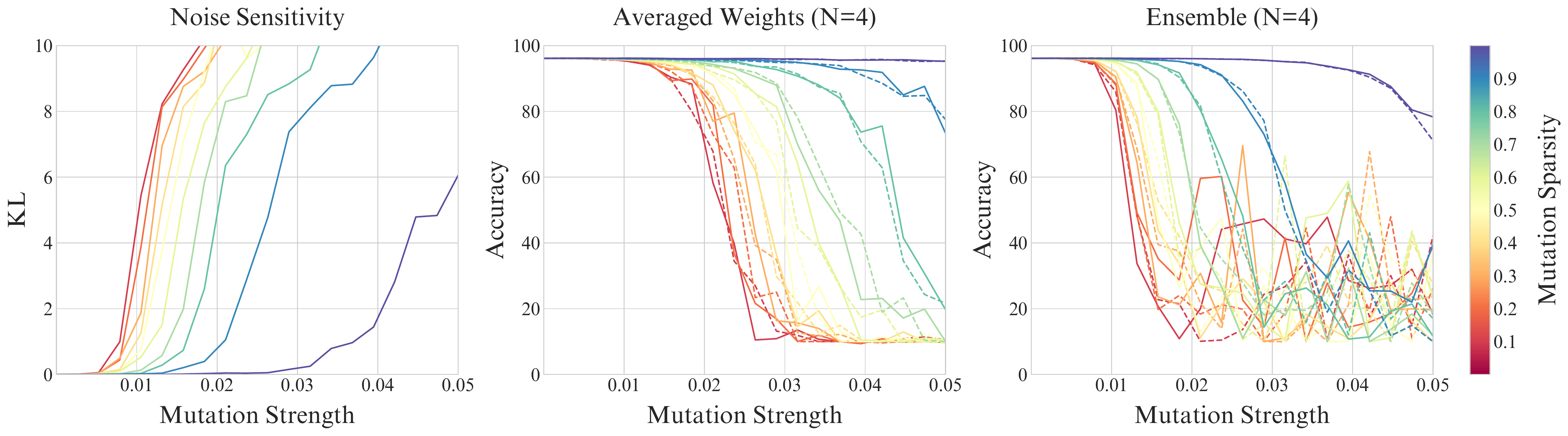}
    \caption{Results of mutation ablations on a trained WideResNet-28x10 model on CIFAR-10. Lines of different colors correspond to the percentage of sparsity for the mutations. Dashed lines correspond to static subspace evolution while solid lines correspond to dynamic subspace evolution. The left most graph reports the average KL divergence. The middle graph reports the accuracy of an averaged model of the four top candidates after one generation of evolution. The rightmost graph reports accuracy if you treat those four models as an ensemble.}
\vspace{0.1in}
\end{figure*}

\begin{table*}
\begin{tabularx}{\textwidth}{X l l l l l l c c c c c}
\toprule
Model & Acc $\uparrow$ & NLL $\downarrow$ & ECE $\downarrow$ & eAcc $\uparrow$ & eNLL $\downarrow$ & eECE $\downarrow$ & $\Delta$Acc $\uparrow$ & Parameters & \multicolumn{1}{c}{$\sigma$} & \multicolumn{1}{c}{$\rho$} & $\overline{KL}$  \\
\midrule
AlexNet & 56.46 & 1.904 & 0.021 & 56.52 & 1.903 & 0.019 & 0.06 & 61.1M & 0.006 & 0.80 & 0.038 \\
\text{DenseNet-121} & 74.43 & 1.014 & 0.024 & 74.68 & 1.004 & 0.021 & 0.25 & 8.0M & 0.007 & 0.85 & 0.077 \\
\text{Inception-V3} & 69.57 & 1.819 & 0.184 & 69.98 & 1.681 & 0.169 & 0.41 & 27.2M & 0.008 & 0.90 & 0.037 \\
\text{MobileNet-V2} & 72.12 & 1.136 & 0.072 & 72.14 & 1.132 & 0.024 & 0.02 & 3.5M & 0.007 & 0.90 & 0.085 \\
\text{ResNet-18}  & 69.76 & 1.247 & 0.026 & 69.93 & 1.238 & 0.022 & 0.17 & 11.7M & 0.010 & 0.90 & 0.060 \\
\text{ResNext-50} & 77.64 & 0.945 & 0.065 & 77.72 & 0.929 & 0.059 & 0.08 & 25.0M & 0.005 & 0.80 & 0.051 \\
\text{ShuffleNet-V2} & 69.51 & 1.354 & 0.072 & 69.57 & 1.351 & 0.071 & 0.06 & 2.3M & 0.012 & 0.95 & 0.052 \\
SqueezeNet & 58.10 & 1.852 & 0.017 & 58.14 & 1.852 & 0.017 & 0.04 & 1.2M & 0.007 & 0.75 & 0.072 \\
\text{VGG-16} & 71.62 & 1.140 & 0.027 & 71.64 & 1.138 & 0.028 & 0.02 & 138.4M & 0.010 & 0.95 & 0.019\\
\text{WideResNet-50} & 78.47 & 0.879 & 0.054 & 78.60 & 0.852 & 0.038 & 0.13 & 68.9M & 0.011 & 0.90 & 0.178 \\
\bottomrule
\end{tabularx}

\small

\caption{Results for sparse mutation ensembles on ImageNet with a wide variety of models. Mutation sparsity and strength are determined from a small hyperparameter grid search. Accuracy (Acc), Negative Log Likelihood (NLL), and expected calibration error (ECE) are reported for the parent and the ensemble. Metrics prepended with $e$ refer to the ensemble results. $\Delta$Acc is the change in accuracy between the parent and the ensemble, $\sigma$ is the mutation strength, $\rho$ is the mutation sparsity, and $\overline{KL}$ is the average output divergence between the mutated models and the parent models. 16 models are generated and the 8 most accurate candidates on a validation set are used together as an ensemble. We see a small but reliable improvement in generalization performance in every single case.}

\end{table*}

Next, we aim to explore whether the intuitions about how mutations affect predictions translate to a much larger convolutional network on benchmark computer vision datasets.
For this experiment, we conduct ablations on the CIFAR-10 and CIFAR-100 datasets \cite{Krizhevsky09learningmultiple}. These are popular benchmark datasets and their use is widespread in computer vision research. They each contain 50,000 training and 10,000 test samples of colored 32x32 pixel images. CIFAR-10 contains samples belonging to one of 10 classes while CIFAR-100 contains samples belonging to one of 100 classes.
We use a WideResNet-28x10 model for our parent network, which is a highly accurate network architecture that contains $\sim 36M$ parameters. This network is a variant on the popular ResNet that decreases the depth and increases the width of each convolutional layer \cite{https://doi.org/10.48550/arxiv.1605.07146}.

We implement a standard training algorithm for this type of model where we train for 100 epochs using Stochastic Gradient Descent with Nestorov momentum \cite{sutskever13nesterov}.
A stepwise learning rate decay is used where an inital value of 0.1 decays to 0.01 after 50\% of training and decays again to 0.001 for the final 10\% of training.
We use standard data augmentation schemes for CIFAR that includes a random crop and random horizontal flip along with mean standard normalization.
We split the test set in half in order to conduct a validation set used for model selection.

The parent networks achieve an accuracy of approximately 96\% on CIFAR-10 and approximately 80\% on CIFAR-100.
Using these saved models as parent networks, we then perturb the models with varying amounts of mutation strengths, $\sigma \in (0.0, 0.05]$, and sparsities, $\rho \in [0.01, 0.99]$, and we evaluate their performance on the test sets.

Figure 3 displays the results of these experiments on CIFAR-10 where we start by measuring the effect that mutations have on KL Divergence.
Predictably, we see that KL divergence quickly increases as the density and strength of mutations increase.
The rapid increase in KL divergence for dense perturbations illustrates how quickly performance collapses between very small changes in mutation strength.

We then implement a single evolutionary generation where a population of 16 models are created by perturbing the parent model according to a given mutation strength and sparsity.
The top 4 models with the best accuracy on the validation set are then selected and evaluated.
We then report accuracy on the test set where the weights of the top 4 models are averaged together.
We also evaluate these four models as if they were part of an ensemble, where each model is evaluated independently and their predictions are combined.
The averaged weights model is much more consistent compared to the ensemble which displays large amounts of variance for strong mutations.
While the differences are small in this case, ensembles tend to slightly outperform averaged weight models for lower levels of mutation strengths (where individual model accuracy is higher) while averaged models outperform ensembles for higher levels of mutation (where individual model performance is worse).

We repeat the above experiments with both static and dynamic subspace evolution.
For static subspace evolution, we mutate the same subnetwork for each generated member in the population. For dynamic subspace evolution, each child is mutated with a random subnetwork mask. On this task, we see little difference between static and dynamic subspaces, suggesting that the specific subnetwork that we mutate does not matter in the context of both averaged and ensemble evaluations.

\subsection{Large Scale Evaluation}

We then explore our approach on the difficult benchmark computer vision dataset, ImageNet \cite{NIPS2012_c399862d}. ImageNet is a large scale collection of images that have been hand labelled for use in machine learning tasks and is organized according to the wordnet dataset hierarchy. Over 14 million images and 20,000 labels have been collected in total. We use the 2012 ImageNet collection which consists of a training set of 1.2 million images and a validation set of 50,000 images, each belonging to one of 1000 categories. Images have varying sizes and aspect ratios and consist of both colored and grayscale photos. We normalize all images are normalized with mean and standard deviation scaling and we implement standard data augmentation which consists of resizing to 256x256 pixels and center cropping to 224x224 pixels.

We evaluate our approach with ten popular deep neural network architectures of varying sizes and generalization capacities in order to demonstrate the generality and power of our approach in many different contexts. All networks are pretrained and available from the torchvision repository \cite{torchvision}. 

We break the ILSVRC 2012 set of 50,000 images into a 50/50 split between validation and test sets that each contain 25,000 samples.
All fitness evaluations and model selections use the validation set and all reported results are evaluated on the holdout test set.

We report accuracy, negative log likelihood, and estimated calibration error for the parent network and our evolutionary ensemble. We run each model twice and report the best results.

We conduct a hyperparameter grid search in order to find appropriate hyperparameter values for mutation strength and sparsity for each model.
Using a separate holdout dataset of 1000 samples we measure the average KL Divergence and accuracy while we ablate sparsity $\rho \in [0.5, 0.99]$ and mutation strength $\sigma \in [0.001, 0.015]$. We then choose both a mutation strength and sparsity value that maximizes accuracy while targeting a KL Divergence of $\sim 0.05$. This value was found to be an empirically safe option for most models, balancing the accuracy of candidate models with exploration to reliably improve generalization performance when candidates are combined.

Using the sparse mutation hyperparameters found from the short grid search, a population of 16 models are created with mirrored sampling.
Each is evaluated on a validation set and the top 8 candidates with the best accuracy are selected.
We then evaluate the top 8 candidates independently and combine their predicted probabilities as if they were part of an ensemble and report the accuracy, negative log likelihood and expected calibration error.

Table I contains both the parent results and the ensemble results. From only one generation of evolutionary tuning, we see consistent improvement in all metrics for each model. While the difference between generalization performance is very small in many cases, it is notable that convergence is extremely stable and appears to be monotonic with optimal mutation hyperparameters. We don't observe the noisy oscillations you'd commonly see with fine tuning using stochastic gradient descent.
There's also a question of how much potential is able to be squeezed out of the fully trained networks. There is no accepted method for determining the theoretical ceiling of generalization capacity for complex network architectures.
It is possible that these small improvements are significant in the grand scheme of things where an improvement of 0.5\% accuracy on a dataset of 50,000 images corresponds to 250 more correct predictions, which can be significant in some contexts.
It is notable that our results only incorporate a single generation of evolutionary fine tuning and future research with more iterations and more thorough hyperparameter searches will likely improve performance further.

\section{Conclusion}

We introduce Sparse Mutation Decompositions as an approach to alleviating the challenges of mutating deep neural networks by breaking up dense mutations into low-dimensional subspaces. This widens the critical mutation window, which can significantly reduce the variance as children in evolutionary populations can handle stronger perturbations before performance collapse. We explore how these sparse mutations can be implemented with a standard neuroevolutionary method in order to fine-tune and further optimize pre-trained networks and we show how ensembles of the top candidates in a population can aid in generalization.

We conduct several ablation studies in order to explore the interplay between sparsity and mutation strength on network behavior. We conduct a decision boundary experiment where visualizations are created that show the predictions of a model on a binary classification dataset. After being perturbed with dense mutations, the model sees a rapid decline in performance while sparsity significantly helps in maintaining accurate predictions while encouraging diverse representation. We then explore how these insights translate to a wide residual network on CIFAR where we explore parameter sensitivity, static/dynamic subspace evolution, and the differences between averaged model performance and population ensemble performance as result of different levels of mutation strength and sparsity. Our findings reaffirm the idea that sparse mutations produce more accurate models more reliably than dense mutations.

We then introduce the first large scale exploration of evolutionary fine tuning with sparse mutations on ImageNet with a wide variety of deep neural network architectures.
We use a relatively small population of 16 models in which the top 8 most accurate on a validation set are selected and evaluated together as an ensemble.
Our approach reliably and consistently improves performance on every model with only a single evolutionary generation.

\bibliographystyle{plain}
\bibliography{bibliography}

\end{document}